# METAHEURISTIC OPTIMIZATION OF POWER AND ENERGY SYSTEMS: UNDERLYING PRINCIPLES AND MAIN ISSUES OF THE 'RUSH TO HEURISTICS'

Gianfranco Chicco and Andrea Mazza

Dipartimento Energia "Galileo Ferraris", Politecnico di Torino
Corso Duca degli Abruzzi 24, 10129 Torino, Italy

**Abstract**

In the power and energy systems area, a progressive increase of literature contributions containing applications of metaheuristic algorithms is occurring. In many cases, these applications are merely aimed at proposing the testing of an existing metaheuristic algorithm on a specific problem, claiming that the proposed method is better than other methods based on weak comparisons. This 'rush to heuristics' does not happen in the evolutionary computation domain, where the rules for setting up rigorous comparisons are stricter, but are typical of the domains of application of the metaheuristics. This paper considers the applications to power and energy systems, and aims at providing a comprehensive view of the main issues concerning the use of metaheuristics for global optimization problems. A set of underlying principles that characterize the metaheuristic algorithms is presented. The customization of metaheuristic algorithms to fit the constraints of specific problems is discussed. Some weaknesses and pitfalls found in literature contributions are identified, and specific guidelines are provided on how to prepare sound contributions on the application of metaheuristic algorithms to specific problems.

## 1. Introduction

Large-scale complex optimization problems, in which the number of variables is high and the structure of the problem contains non-linearities and multiple local optima, are computationally challenging to solve, as they would require computational time beyond reasonable limits, and/or excessive memory with respect to the available computing capabilities. These problems appear in many domains, with different characteristics. A number of optimization problems are characterized by objective functions with non-smooth surfaces corresponding to the solution points, presence of discrete variables, and several local minima, as well as the combinatorial explosion of the number of cases to be evaluated to search for the global optimum. The algorithms used to solve these problems have to be able to perform efficient *global optimization*.

In general, the solution algorithms are based on two types of methods:
1) *deterministic methods*, in which the solution strategy is driven by well-identified rules, with *no random* components; and,
2) *probability-based methods*, whose evolution depends on *random* choices carried out during the evolution of the solution process.

If the nature and size of the problem enable convenient formulations (e.g., linearizing the non-linear components by means of piecewise linear representations in a convex problem structure), some *exact* deterministic methods can reach the global optimum under specific data representations. However, in general the structure of the problems can be so complex to make it impossible or impracticable to use methods that can guarantee to reach the global optimum. Furthermore, in some cases, even the structure of the solution space is unknown, thus needing a specific approach to obtain information from the solutions themselves. In these cases, a viable approach is the use of metaheuristics.

What is a metaheuristic? In synthesis, the term *heuristic* identifies a tool that helps us discover 'something'. The term *meta* is typically added to represent the presence of a higher-level strategy that drives the search of the solutions. The metaheuristics could depend on the specific problem (Sörensen et al. 2018). Many metaheuristics are based on translating the representation of *natural* phenomena or *physical processes* into computational tools (Salcedo-Sanz 2016).

Some solution methods are *metaheuristics* based on one of the following mechanisms:

a) *Single solution update*: a succession of solutions is calculated, each time updating the solution only if the new one satisfies a predefined criterion. These methods are also called *trajectory* methods.
   b) *Population-based search*: many entities are simultaneously sent in parallel to solve the same problem. Then, the collective behavior can be modeled to link the different entities with each other, and in general the best solution is maintained for the next phase of the search.

Detailed surveys of single solution-based and population-based metaheuristics are presented (among others) in Boussaïd et al. (2013), Zedadra et al. (2018), and Dorekoglu et al. (2019).

On another point of view, considering the number of optimization objectives, a distinction can be indicated among:
   (i) *Single objective* optimization, in which there is only one objective to be minimized or maximized.
   (ii) *Multi-objective* optimization, in which there are two or more objectives to be minimized or maximized. Multi-objective optimization tools are significant to assist decision-making processes, when the objectives are conflicting with each other. In this case, an approach based on Pareto-dominance concepts becomes useful. In this approach, a solution is *non-dominated* when no other solution does exist with better values for all the individual objective functions. The set of non-dominated solutions forms the *Pareto front*, which contains the compromise solutions among which the decision-maker can choose the preferred one. If the Pareto front is convex, the weighted sum of the objectives can be used to track the compromise solutions. However, in general, the Pareto front has a non-convex shape, calling for appropriate solvers to construct it. During the solution process, the best-known Pareto front is updated until a specific stop criterion is satisfied. The best-known Pareto front should converge to the true Pareto front (that could be unknown). The solvers need to balance *convergence* (i.e., approaching a stable Pareto front) with *diversity* (i.e., keeping the solutions spread along the Pareto front, avoiding to concentrate the solutions in limited zones). Diversity is represented by estimating the density of the solutions located around a given solution. For this purpose, a dedicated parameter called *crowding distance* is defined as the average distance between the given solution and the closest solutions belonging to the Pareto front (the number of solutions is user-defined).
   (iii) *Many-objective* optimization, a subset of multi-objective optimization in which conventionally there are more than two objectives. This distinction is important, as some problems that are reasonably solvable in two dimensions, such as finding a balance between convergence and diversity, become much harder to solve in more than two dimensions. Moreover, by increasing the number of objectives, it becomes intrinsically more difficult to visualize the solutions in a way convenient for the decision-maker. The main challenges in many-objective optimization are summarized in Li et al. (2015). When the number of objectives increases, the number of non-dominated solutions largely increases, even reaching situations in which almost all solutions become non-dominated (Ishibuchi et al. 2008). This aspect heavily impacts on slowing down the solution process in methods that use Pareto dominance as a criterion to select the solutions. A large number of non-dominated solutions may also require to increase the size of the population to be used in the solution method, which again results in a slower solution process. Finally, the calculation of the hyper-volume as a metric for comparing the effectiveness of the Pareto front construction from different methods (Zitzler & Thiele 1999), is geometrically simple in two dimensions, but becomes progressively harder (Guerreiro & Fonseca 2018), with the exponential growth of the computational burden, when the number of objectives increases (While et al. 2006).

For multi-objectives and many-objectives, the metaheuristic approach has gained momentum because of the issues existing in the application of gradient search and numerical programming methods. However, a number of issues appear, mainly concerning the characteristics of the search space (such as non-convexity and multimodality), and the presence of discrete non-uniform Pareto fronts (Zitzler et al. 2000).

The above indications set up the framework of analysis used in this paper. The main aims are to start from the concepts referring to the formulation of the metaheuristics, and discuss a number of correct and inappropriate practices found in the literature. Some details are provided on the applications in the power and energy systems domain, in which hundreds of papers based on the use of metaheuristics for solving optimization problems have been published.

The specific contributions of this paper are:
- A systematic analysis of the state of the art concerning the main issues on the convergence of metaheuristics, and the comparisons among metaheuristic algorithms with suitable metrics for single-objective and multi-objective optimization problems.

- The identification of a set of underlying principles that explain the characteristics of the various metaheuristics, and that can be used to search for similarities and complementarities in the definition of the metaheuristic algorithms.
- The indication of some pitfalls and inappropriate statements sometimes found in literature contributions on global optimization through metaheuristic algorithms, which contribute to the proliferation of articles on the application of metaheuristics not always justified by a rigorous methodological approach, leading to an almost uncontrollable 'rush to heuristics'.
- The discussion on the characteristics of some problems in the power and energy system domain, solved with metaheuristic optimization, highlighting some problem-related customizations of the classical versions of the metaheuristic algorithms.
- The indication of some hints for preparing sound contributions on the application of metaheuristic algorithms to power and energy system problems with one or more objective functions, using statistically significant and sufficiently strong metrics for comparing the solutions, in such a way to mitigate the 'rush to heuristics'.

The next sections of this paper are organized as follows. Section 2 summarizes the underlying principles that can be found in the construction of metaheuristic algorithms. Section 3 recalls the convergence properties of some metaheuristics. Section 4 deals with the hybridization of the metaheuristics. Section 5 addresses the use of metaheuristics to solve multi-objective problems. Section 6 discusses the effectiveness of metaheuristic-based optimization, pointing out a number of weak statements that should not appear in scientific contributions. The last section contains the concluding remarks.

## 2. Underlying principles for representing the characteristics of the metaheuristics

The evolution of the metaheuristics has increased progressively in time. New algorithms appear each year, and it is not clear-cut whether these algorithms bring new contents for the research on evolutionary computation. Table 1 shows a non-exhaustive list of over one hundred metaheuristics that have been applied in the power and energy systems field. The years indicated refer to the first date of publication of relevant articles or books. Figure 1 shows the corresponding number of metaheuristics used during the time. The number of metaheuristics that appeared in the last years is underestimated, as some recent metaheuristics (not indicated) have not found an application in power and energy systems yet. Moreover, the list in Table 1 refers to basic versions of the metaheuristics only, without accounting for the proposed variants and hybridizations among heuristics; otherwise, the number of contributions would quickly rise to significantly higher numbers. However, the rush to apply a new metaheuristic to all the engineering problems is a vulnerable point for scientific research (Sörensen 2015), especially when each "new" method or variant applied to a given problem is claimed to become the best method, pretending to show its superiority with respect to any other existing method. Apparently, this 'rush to heuristics' is producing hundreds of articles, most of them questionable in terms of methodological advances provided in the evolutionary computation field.

The need to better understand the characteristics of the various algorithms has started specific discussions since the early phase of the development of new algorithms. Two decades ago, the unified view proposed in Taillard et al. (2001) started from the consideration that the implementation of the solvers was increasingly similar. The unified view was introduced under the name Adaptive Memory Programming (AMP), synthesizing a series of basic steps for the solution procedure valid for most metaheuristics (AMP is not applicable to single-update methods such as Simulated Annealing (Kirkpatrick et al. 1983)):

1) Store a set of solutions
2) Construct a provisional solution using the available data
3) Improve the provisional solution with *local search* or another algorithm
4) Update the set of available solutions with the new solution

These steps indicate four basic principles used to set up a metaheuristic algorithm, namely, memory (i.e., storage of information), the presence of a constructive mechanism, a local search strategy, and the definition of a mechanism for solution update. Taillard et al. (2001) consider *memory* as a key principle for describing the possible similarities between the algorithmic structures of the metaheuristics. Indeed, memory is fundamental in the definition of the metaheuristics. However, memory can be seen as a general term with different meanings for different algorithms. As such, memory is not considered here to be sufficiently specific as a detailed underlying principle, and more underlying principles are used for representing the characteristics of the various types of metaheuristics.

Table 1. Over one hundred heuristics used in the power and energy systems domain.

| Heuristic | Year | Heuristic | Year |
|---|---|---|---|
| Ant colony optimization | 1991 | Group search optimization | 2006 |
| Ant-lion optimizer | 2015 | Harmony search algorithm | 2013 |
| Artificial algae algorithm | 2015 | Harris hawks optimizer | 2019 |
| Artificial ecosystem-based optimization | 2019 | Imperialist competitive algorithm | 2007 |
| Artificial bee colony | 2007 | Intelligent water drops | 2007 |
| Artificial cooperative search algorithm | 2013 | Invasive weed optimization | 2006 |
| Artificial fish swarm algorithm | 2018 | Ions motion optimization algorithm | 2015 |
| Artificial immune system | 1986 | Jaya algorithm | 2016 |
| Atom search optimization | 2019 | Kinetic gas molecule optimization | 2014 |
| Auction-based algorithm | 2014 | Krill herd algorithm | 2012 |
| Backtracking search algorithm | 2013 | League championship algorithm | 2014 |
| Bacterial foraging | 2002 | Lion optimization algorithm | 2016 |
| Bat-inspired algorithm | 2010 | Manta ray foraging optimization | 2020 |
| Bayesian optimization algorithm | 1999 | Marine predators algorithm | 2020 |
| Big-bang big-crunch | 2013 | Marriage in honey bees optimization | 2001 |
| Biogeography based optimization | 2011 | Mean-variance mapping optimization | 2010 |
| Brainstorming process algorithm | 2011 | Melody search algorithm | 2013 |
| Cat swarm optimization | 2006 | Memetic algorithms | 1989 |
| Chaos optimal algorithm | 2010 | Mine blast algorithm | 2013 |
| Charged system search | 2010 | Monarch butterfly optimization | 2015 |
| Chemical reaction based optimization | 2010 | Monkey algorithm | 2007 |
| Civilized swarm optimization | 2003 | Moth-flame optimization | 2015 |
| Clonal selection algorithm - Clonalg | 2002 | Optics inspired optimization | 2014 |
| Cohort Intelligence | 2013 | Particle swarm optimization | 1995 |
| Coral reefs optimization | 2014 | Pigeon inspired optimization | 2014 |
| Covariance matrix adaptation evolution strategy | 2003 | Population extremal optimization | 2001 |
| Colliding bodies optimization | 2014 | Plant growth simulation | 2005 |
| Coyote optimization algorithm | 2018 | Predator–prey optimization | 2006 |
| Crisscross optimization algorithm | 2014 | Quantum-inspired evolutionary algorithm | 1995 |
| Crow search algorithm | 2016 | Quick group search optimizer | 2010 |
| Cuckoo search algorithm | 2009 | Radial movement optimization | 2014 |
| Cultural algorithm | 1994 | Rain-fall optimization | 2017 |
| Dendritic cell algorithm | 2005 | Ray optimization algorithm | 2012 |
| Differential evolution | 1997 | River formation dynamics | 2007 |
| Differential search algorithm | 2013 | Salp swarm algorithm | 2017 |
| Diffusion limited aggregation | 1981 | Simulated annealing | 1983 |
| Dolphin echolocation algorithm | 2013 | Scatter search | 1977 |

| Heuristic | Year | Heuristic | Year |
|---|---|---|---|
| Dragonfly algorithm | 2016 | Seagull optimization | 2019 |
| Eagle strategy | 2010 | Seeker optimization algorithm | 2006 |
| Electromagnetism-like mechanism | 2012 | Shuffled frog leaping algorithm | 2006 |
| Election algorithm | 2015 | Sine-cosine algorithm | 2016 |
| Elephant herd optimization | 2015 | Slime mould optimization algorithm | 2008 |
| Equilibrium optimizer | 2020 | Soccer league competition algorithm | 2014 |
| Estimation of distribution algorithms | 1996 | Social group optimization | 2016 |
| Evolutionary algorithms | 1966 | Social spider algorithm | 2015 |
| Evolution strategies | 1971 | Squirrel search algorithm | 2019 |
| Farmland fertility optimization | 2018 | Stochastic fractal search | 2015 |
| Firefly algorithm | 2010 | Symbiotic organisms search | 2014 |
| Firework algorithm | 2010 | Tabu search (*) | 1989 |
| Flower pollination algorithm | 2012 | Teaching-learning-based optimization | 2011 |
| Front-based yin-yang-pair optimization | 2016 | Tree-seed algorithm | 2015 |
| Fruit fly optimization | 2012 | Variable neighborhood search | 1997 |
| Galactic swarm optimization | 2016 | Virus colony search | 2016 |
| Galaxy-based search algorithm | 2011 | Volleyball premier league | 2018 |
| Gases Brownian motion | 2013 | Vortex search algorithm | 2015 |
| Genetic algorithms | 1975 | Water cycle algorithm | 2012 |
| Glowworm swarm optimization | 2005 | Water waves optimization | 2015 |
| Grasshoppers optimization | 2017 | Weighted superposition attraction | 2016 |
| Gravitational search algorithm | 2009 | Whale optimization algorithm | 2016 |
| Greedy randomized adaptive search procedures | 1989 | Wind driven optimization | 2010 |
| Grenade explosion method | 2010 | Wolf search algorithm | 2012 |
| Grey wolf optimization | 2014 | (*) not based on random number extractions | |

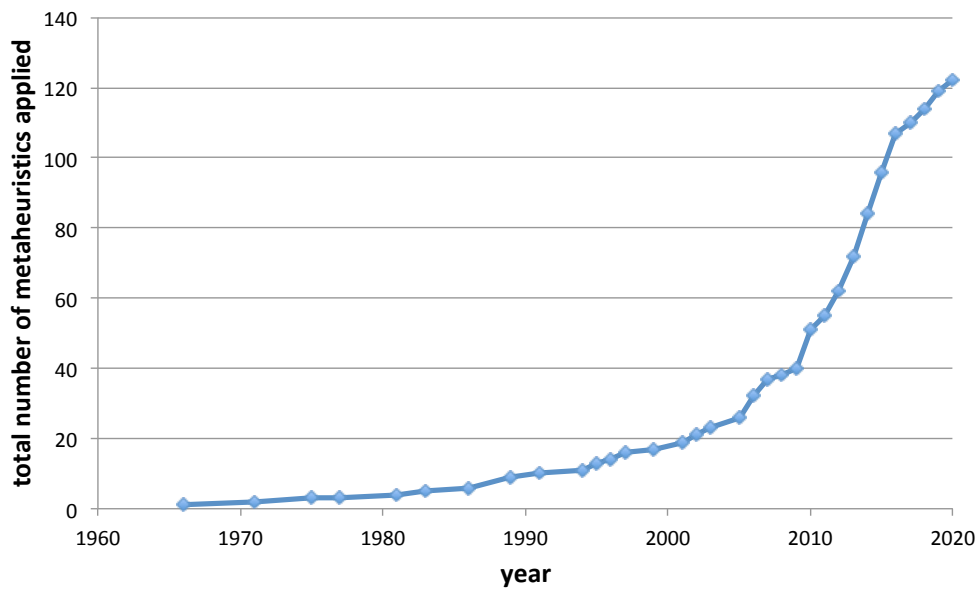

Figure 1. Number of metaheuristics available (variants and hybrid versions excluded).

*2.1. Underlying principles*

Each metaheuristic algorithm applies specific mechanisms in the solution procedure. The presence of a multitude of algorithms raises a fundamental question: are all the metaheuristic algorithms used really different from each other?

To address this issue, the solution procedures have been revisited by identifying a set of underlying principles that form a common basis for the various methods (Batrinu et al. 2005). On the other side, these principles embed the structural differences among the methods.

The following list of underlying principles has been found, which also highlights some contents referring to typical issues that appear in the power and energy systems domain:

- Parallelism
- Acceptance
- Elitism
- Selection
- Decay (or reinforcement)
- Immunity
- Self-adaptation
- Topology

A brief description of these principles follows.

*2.2. Parallelism*

The parallelism principle appears in population-based search, in which more entities are sent in parallel to perform the same task, and the results obtained are then compared. On the basis of the comparison, further principles are applied to determine the evolution of the individuals within the population or to create new populations.

*2.3. Acceptance*

The principle of acceptance appears in a threefold way:
1. Temporarily accept solutions that lead to objective function *worsening*, with the rationale of broadening the search space
2. In the treatment of the *constraints* applied to the objective function. The constraints can be handled in two different ways. The first way is to discard all solutions in which any violation appears. This way is applied to algorithms that use a non-penalized objective function, in which the initial conditions have to correspond to a feasible solution (for single-update methods) or to all feasible solutions (in population-based methods). The second way is to use a penalized objective function, which makes it possible to find out a numerical value to any solution and avoid discarding any solution. In this case, all solutions are automatically accepted, and the initial conditions could correspond to infeasible solutions. The penalty factors used in the penalized objective functions have to be sufficiently high to obtain high values for the solutions with violations. However, if the penalty factor is too high, very high values could appear for too many solutions, making it difficult to drive the search in the direction of exploring the search space efficiently.
3. Introducing a *threshold* for accepting only solutions that improve the current best solution at least of the value of the threshold. This way could help to avoid numerical issues in the comparison between values resulting from previous calculations, e.g., when the same number is represented in different ways depending on numerical precisions.

*2.4. Elitism*

In the iterative population-based methods (in which more individuals are generated at the same iteration from probability-based criteria), if no action is done, it is possible to lose the best solution passing from one iteration to another. The basic versions of metaheuristics (such as simulated annealing, genetic algorithms, and others), privilege the randomness of the search and do not contain mechanisms to preserve the best solutions. To avoid this, the *elitism* principle is applied by storing the individual with the best objective function found so far and passing it from one iteration to the next one. The best solution can be used as a reference individual to form other modified solutions, and is immediately updated when another best solution is found. In a more extensive way, the elitism principle can also be applied to more than one individual, passing an *élite group* of solutions to the next iteration. The elitism principle has resulted very effective in practical applications. For the elitistic versions of some metaheuristics it has been possible to prove convergence to the global optimum under specified conditions (see Section 3).

*2.5. Selection*

In a probability-based method, a mechanism has to be identified to extract a number of individuals at random from an available population, possibly associating weights to the probabilistic choices. In particular, for problems with variables described in a discrete way, the extraction mechanism is driven by the conventional way to extract a point from a given probability distribution. The Cumulative Distribution Function (CDF) is constructed by considering a quality measure (fitness) of the solutions, reported in a normalized way, such as the individuals corresponding to better values of the objective function have higher fitness (hence higher probability to be chosen). For example, let us consider a set of *M* individuals, whose objective function values are $\{v_m > 0, m = 1, \ldots, M\}$, and the objective function has to be minimized (Carpaneto & Chicco 2008). The fitness is defined as

$$\psi_m = \frac{v_m}{\sum_{q=1}^{M} v_q} \tag{1}$$

Then, a random number *r* is extracted from a uniform probability distribution in [0, 1] and is entered on the vertical axis of the CDF. The individual corresponding to the discrete position on the *horizontal* axis is then selected. The situation is exemplified in Figure 2, with four individuals and the related fitness values 0,2, 0.3, 0,4, and 0.1, respectively. By extracting a random number (e.g., 0.62), the individual number 3 is selected.

This method is equivalent to the so-called biased roulette wheel, in which the slices have a different amplitude (proportional to the fitness associated to each individual). The selected variable is the one seen from the observation at which the roulette stops. In the example of Figure 3, the individual D has the largest probability of being selected, but *any* individual can be selected (e.g., the individual H is selected in Figure 3).

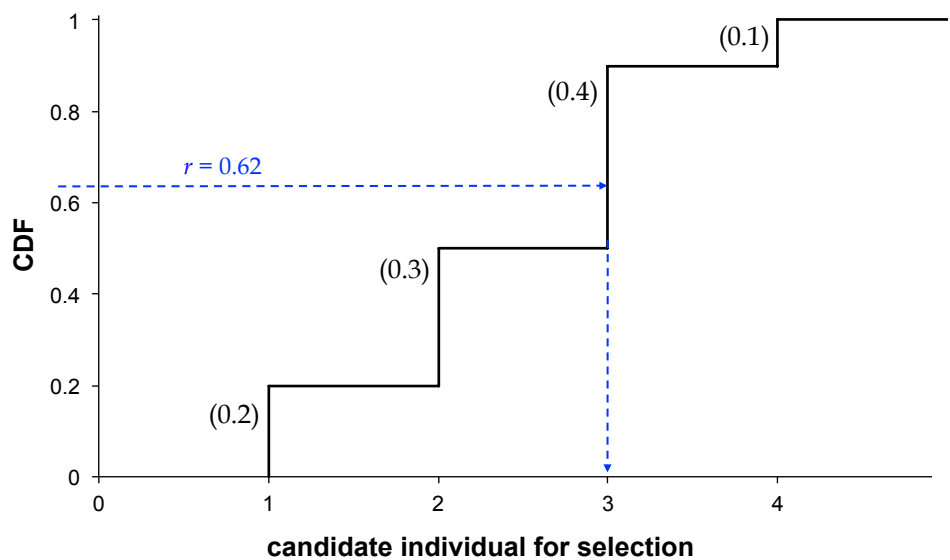

Figure 2. Random selection from CDF.

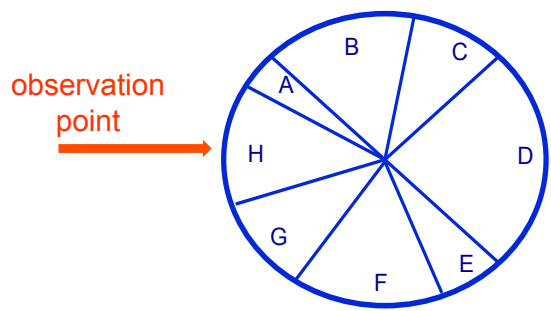

Figure 3. Random selection from biased roulette wheel.

*2.6. Decay (or reinforcement)*

The decay principle may be applied to enable larger initial flexibility in the application of the method, followed by progressive restrictions of that flexibility. The application of a decay rate to the parameter (cooling rate) that drives the external cycle in the simulated annealing method is a direct example. Decay is typically considered by using a multiplicative factor lower than unity, which is applied at successive iterations. In some cases, reinforcement is applied in a similar way by using a multiplicative factor higher than unity.

The decay principle can also reduce the strength of some search paths that are less convenient than others or have not been visited recently. This application has been introduced in the ant colony optimization algorithms (Dorigo et al. 1991), in which the paths can also be reinforced if they were found to be convenient. Relative decay has also been considered in the hyper-cube ant colony optimization framework (Blum & Dorigo, 2004), in which decay or reinforcement are first applied, then the overall outcomes are normalized to fit an hyper-cube with dimensions limited inside the interval [0, 1].

*2.7. Immunity*

Immunity is applied by identifying some properties of the solutions, where such properties lead to satisfactory configurations. Immunity gives *priority* to the solutions having characteristics similar to those properties.

*2.8. Self-adaptation*

Self-adaptation consists of changing the parameters of the algorithms in an automatic way, depending on the evolution of the procedure.

*2.9. Topology*

The topology principle is applied when the problem under analysis needs to satisfy specific constraints such as the definition on a graph, or connectivity requirements. A relevant example is the *graph* corresponding to the operational configuration of an electrical distribution system. The principle of topology is linked, for example, to the generation of radial structures during the execution of the algorithms. The representation of the topology is associated with how the information regarding the connections is coded, which can be more or less effective to ensure that only radial structures are progressively generated. For example, for an electrical network, the information coding is typically carried out in one of these ways:
   a) creating the list of the open branches;
   b) forming the list of the loops and identifying the branches of each loop with a progressive number; or,
   c) using a *binary string* of length equal to the number of branches, containing the status (on/off) of the branches.

*2.10. Remarks on the underlying principles*

The identification of the underlying principles has clarified that the memory term can be intended in different ways, e.g., pheromone for ant colony optimization, the presence of the previous population for genetic algorithms, the list of past moves for tabu search, and so on. Elitism itself is a form of memory.

In the use of metaheuristics, a balance is generally sought between exploration and exploitation of the search space. Exploration means the ability to reach all the points in the search space, while exploitation refers to the usage of knowledge from the solutions already found to drive the search towards more convenient regions. The underlying principles may affect both exploration and exploitation in different ways. For example, the selection principle applied to a given population could mainly refer to exploitation (Chen et al. 2009), as it drives the search towards the choice of the best individuals (Črepinšek et al. 2013). However, selection may to a given extent refer to exploration as well, by varying the width of the population involved (Bäck 1994).

Finally, the synthesis of the underlying principles can also be a way to generate new metaheuristic algorithms or variants. Even automatic generation of algorithms could be considered, for which there is a wide literature referring to deterministic and other algorithms (Mitsos et al. 2018). Indeed, conceptually there are different ways to proceed to define new metaheuristic algorithms:
   a) Taking existing algorithms and constructing new ones by changing the context and the nomenclature; this practice is indeed not advancing the state of the art, and only contributes to add entropy to the evolutionary computation domain (Sörensen 2015).

b) Synthesizing the underlying principles and combining them to obtain new algorithms; also in this case, it is only a recombination of existing principles, which in general could not add significant contributions and would just play into the 'rush to heuristics'.
c) Generate new algorithms by using a set of components taken from promising approaches (Bain et al. 2004). This line of research is also useful to identify appropriate reusable portions of subprograms (Koza 1994), and also leads to using *hyper-heuristics* to select or generate (meta-)heuristics by exploring a search space of a number of heuristics for identifying the most effective ones (Burke et al. 2010; Drake et al. 2020).

Some useful variants can be found when the existing metaheuristics are customized to solve specific problems, incorporating specific constraints, as indicated in the next section.

*2.11. Specific problems for power and energy systems*

In the power and energy domain, metaheuristic optimization is widely used to solve many problems referring to operation, planning, control, forecasting, reliability, security and demand management. A set of typical problems solved with metaheuristic optimization have been considered in Lee et al. (2008) and in Chicco & Mazza (2019), including distribution system reconfiguration, economic dispatch, load forecasting, maintenance scheduling, optimal power flow, and power system planning. A selection of the metaheuristics most applied to typical problems in the specific domain has been shown. From this set of problems, it emerges that the genetic algorithm is the most used or mentioned method for all the problems, followed by particle swarm optimization (or simulated annealing for maintenance scheduling). A review of the particle swarm optimization applications to power systems is presented in del Valle et al. (2008).

Concerning the information coding, the most successful implementations of genetic algorithms do not use binary coding of the strings, but use representations adapted to the application, and the crossover and mutation operators are re-defined accordingly (Taillard et al. 2001). However, binary coding schemes are still mainly used in power and energy system problems. In alternative, evolutionary programming schemes, in which the binary values are replaced with integer or real numbers, are appropriate for specific problems.

Two specific examples are presented below, to indicate how metaheuristics may be a viable alternative to (or a more successful option than) mathematical programming tools for the solution of some large-scale optimization problems in the power and energy systems area. A general remark is that the size of the problem matters. If the size of the problem is limited, for which exhaustive search could be practicable, or mathematic programming tools able to provide exact solutions can be used with reasonable computational burden, then the use of a metaheuristic algorithm is not justified.

*2.11.1. Unit commitment*

The problem consists of scheduling the generation units (typically thermal units) to serve the forecast demand in future periods (e.g., from one day to one week), by minimizing the total generation costs. The output is the start-up and shut-down schedule of these generation units. The problem has integer and continuous variables, and a complex set of constraints also involving time-dependent constraints for the units, such as minimum up and down times, start-up ramps, and time-dependent start-up costs. The unit commitment problem has been traditionally solved with mathematical programming and stochastic programming tools (Zheng et al. 2015). However, these tools exhibit some drawbacks. For example, for dynamic programming, the computation time could become prohibitive for real-size systems, and time-dependent constraints are hard to be successfully implemented. Lagrangian relaxation has no problem with time-dependent constraints and optimizes each unit separately. Thus the dimension of the system is not an issue. The problem is solved by using duality theory, maximizing the dual objective function for a given original problem. However, because of the non-convexity of the original problem, the solution of the dual problem cannot guarantee the feasibility of the primal problem, and optimal values for the original and dual problems could be different. Robust optimizations with bi-level or three-level are computationally less demanding than stochastic programming models, but may lead to over-conservative solutions. A framework for comparing mathematic programming algorithms to solve the unit commitment problem has been formulated in Tejada-Arango et al. (2020) and has been applied to three recently developed algorithms.

To overcome these difficulties, metaheuristics have been successfully used to solve the unit commitment problem. First of all, the binary coding common to various metaheuristics is fully appropriate to represent the on/off status of the units. Thereby, the information on the status of each unit is included in a binary string with length equal to the number of time intervals considered. This information coding is naturally leading to the use of genetic algorithms (Kazarlis et al. 1996), in which a unique string (called chromosome) is constructed as the ordered succession of the strings referring to the individual generation units. The whole information on the scheduling is then available at each time. However, Kazarlis et al. (1996) showed that the straightforward application of the basic version of the genetic algorithm does not lead to acceptable performance, requiring the addition of specific problem-related operators to significantly enhance the algorithm performance. The need to define specific operators has also been confirmed in next implementations, for example, in Swarup & Yamashiro (2002). Further review on the application of metaheuristics to the unit commitment problem is presented in Muralikrishnan et al. (2020).

*2.11.2. Distribution system reconfiguration*

The problem concerns the selection of the open/closed network branches to optimize a predefined objective (or multi-objective) function. The constraints refer to the need to operate a *radial* network, together with the equality constraint on the power balance and inequality constraints involving node voltages, branch currents, short circuit currents, and others (Carpaneto & Chicco 2008). In this case, the discrete variables are the open/closed states of the network branches (or switches, with two switches for each branch, located at the branch terminals). The binary coding of the information is easily applicable. The length of the string is equal to the number of branches (or switches). Additional information, such as the branch list, can be added (Tomoiaga et al. 2013). The number of the possible radial configurations in real-size systems is too high to allow the construction of all radial configurations (Andrei & Chicco 2008). The structure of the problem makes it difficult to identify a neighborhood of the solutions and other regularities that could drive mathematical programming approaches. As such, metaheuristic algorithms are viable to approach this problem.

The main issues for the distribution system reconfiguration problem refer to the implementation of the constraints. In particular, the radiality constraint is not always easily incorporated in the metaheuristic algorithm. For some algorithms, such as simulated annealing, it is sufficient to exploit the branch-exchange mechanism, that consists of starting from the list of the open branches, closing (at random) an open branch to close (that will form a loop), identify the loop, and choosing (at random) within the loop a closed branch to open. In this way, the radial structure of the network is automatically guaranteed. However, for a genetic algorithm, the application of the crossover and mutation operators is not consistent with keeping the radial network structure. Hence, the crossover and mutation operators, and the information coding itself (Carreno et al. 2008) have to be suitably re-defined to ensure radiality is not lost.

## 3. Convergence aspects of global optimization problems and metaheuristics

Metaheuristic algorithms are applied to solve global optimization problems also when the problem structure is not known. For exploring the search space, these algorithms are generally based on the use of random variables, which make it possible to follow non-deterministic paths to reach a solution. The basic versions of the metaheuristic algorithms are relatively simple to be implemented, even though their customization to engineering problems could be very challenging. The metaheuristic algorithms are counterparts of stochastic methods such as two-phase methods, random search methods, and random function methods (Pardalos 2000). In the two-phase methods, the objective function is assessed in a number of points selected at random. Then a local search is carried out to refine the solutions starting from these points. In the random search methods, a sequence of points is generated in the search space by considering some probability distributions, without following with local search. In the random function methods, a stochastic process consistent with the properties of the objective function has to be determined. With respect to these methods, the metaheuristic approach adds a high-level strategy that drives the solutions according to a specific rationale. However, the key point for confirming the significance of metaheuristics is the possibility of proving their convergence in a rigorous way. On the mathematical point of view, convergence proofs are not established for all metaheuristics. Two examples are provided:
1) *Genetic algorithms*: following the introduction of the concepts of genetic algorithms in Holland (1962), the canonical genetic algorithm shown in Holland (1975) did not preserve the best solutions during the evolution of the algorithm, namely, the elitism principle was not applied. In the homogeneous version of the canonical genetic algorithm, the crossover and mutation probabilities always remain constant. For this homogeneous canonical genetic algorithm, there is no proof of convergence to the global optimum. However, better results have been obtained under the condition of ensuring the survival of the best individual with probability equal to unity (elitist selection). In this case, finite Markov chain analysis has been used to prove probabilistic convergence to the best solution in Eiben et al. (1991). The proof that the *elitist* homogeneous canonical genetic algorithm converges *almost surely* to a population which has an optimum point in it has been given in Rudolph (1994). Then, a number of conditions to ensure asymptotic convergence of genetic algorithms to the global optimum have been given in Cerf (1998). Conceptually, due to the application of the genetic operators, at each generation, there is a non-zero probability that a new individual reaches the global optimum. As such, saving the best individual at each generation (in the elitist version) and running the algorithm for an infinite number of generations guarantees that the global optimum can be reached. Further indications to extend the proof of almost sure convergence to the elitist non-homogeneous canonical genetic algorithm are provided in Rojas Cruz et al. (2013), by considering that the mutation and crossover probabilities are allowed to change during the evolution of the algorithm (Campos et al. 2013).

2) *Simulated annealing*: a proof of convergence has been given in Bélisle (1992) for a particular class of algorithms, and the asymptotic convergence has been proven for the algorithm shown in Romeijn & Smith (1994). Further results have been indicated in Locatelli (1996), showing convergence to the global optimum for continuous global optimization problems under specific conditions for the cooling schedule, the function under analysis, and the feasible set.

For multiobjective optimization problems, the proofs of convergence have been set up by introducing elitism, following the successful practice found for single-objective functions. For some multi-objective evolutionary algorithms, convergence proofs to the global optimum are provided in Rudolph (1998) and Rudolph & Agapie (2000). The asymptotic convergence analysis of Simulated Annealing, an Artificial Immune System and a General Evolutionary Algorithm (with any algorithm in which the transition probabilities use a uniform mutation rule) for multiobjective optimization problems is shown in Villalobos-Arias et al. (2005).

**4. Comparisons among metaheuristic algorithms**

*4.1. No free lunches?*

Comparing different algorithms is a very challenging task. Unfortunately, many articles concerning metaheuristics applications in the power and energy systems area (as well as in other engineering fields) are underestimating the importance of this task, and propose simplistic comparison criteria and metrics, such as the best solution obtained, the evolution in time of the objective function improvement for a single run, or related criteria.

In the literature, there is a wide discussion on the algorithm comparison aspects. One of the contributions that have opened an interesting debate is the one that introduced the No Free Lunch (NFL) theorem(s) (Wolpert & Macready, 1997). These theorems state that "any two optimization algorithms are equivalent when their performance is averaged across all possible problems" (Wolpert & Macready, 2005). Basically, the NFL theorems state that no optimization algorithm results in the best solutions for all problems. In other words, if a given algorithm performs better than another on a certain number of problems, there should be a comparable number of problems in which the other algorithm outperforms the first one. However, if a given problem is considered, with its objective functions and constraints, some algorithms could perform better than others, especially when these algorithms are able to incorporate specific knowledge on the problem at hand. The debate includes contributions that argue that the NFL theorems are of little relevance for the machine learning research (Giraud-Carrier & Provost 2005), in which meta-learning can be used to gain experience on the performance of a number of applications of a learning system.

*4.2. Comparisons among metaheuristics*

A recent contribution (Liu et al. 2020) has addressed comparison strategies and their mathematical properties in a systematic way. The numerical comparison between optimization algorithms consists of the selection of a set of algorithms and problems, the testing of the algorithms on the problems, the identification of comparison strategy, methods and metrics, the analysis of the outcomes obtained from applying the metrics, and the final determination of the results.

One of the main issues for setting up the comparisons is the definition of the overall scenario in which the comparison is carried out. The use of benchmarking methodologies such as Black-Box Optimization Benchmarking (BBOB) discussed in Mersmann et al. (2015) pointed out that reaching consensus on ranking the results from evaluations of individual problems is a crucial issue. It is then hard to provide a response to the question "which is the best algorithm to solve a given problem?" (Bartz-Beielstein et al. 2010). However, in some cases, a response should be given, as in the case of competitions launched among algorithms.

When testing a single (existing or new) algorithm, typically a set of algorithms that provide good results for similar problems are selected to carry out the comparison. This is one of the weak points that are encountered in the literature, especially when the choice of the benchmark problems is carried out by the authors without a clear and convincing criterion. A number of mathematical functions that can be used as standard benchmarks are available (Salcedo-Sanz 2016). Some test problems have been defined in different contexts (Gaviano et al. 2003; Hansen et al. 2010; Liang et al. 2013). However, a systematic guide on how to select the set of problems is still missing. The hint given in Liu et al. (2020) is to select the whole set of optimization problems in a given domain, and not only a partial set.

Moreover, for global optimization, there is no known mathematical optimal condition to be satisfied for stopping the search for all the problems. Thereby, the computation time is generally taken as the common limit for stopping the algorithms. For deterministic algorithms, a typical comparison metric is the performance ratio (Dolan & Moré 2002), namely, the ratio between the computation time of the algorithm and the minimum computation time of all algorithms applied to the same problem), from which the performance profile is obtained as the CDF of the performance ratio. Furthermore, the data profile (Moré & Wild 2009) is based on the CDF of the problems that can be solved (by reaching at least a certain target in the solution) with a number of function evaluations not higher than a given limit. With non-deterministic algorithms, the concepts used in the definition of performance profiles and data profiles could be exploited. Comparisons can be carried out by implementing all algorithms on the same computer and running them for the same computational time. The quality of the algorithms can then be determined by calculating the percentage of the best solutions, averaged over a given number of executions of each algorithm (Taillard at al. 2001). The series of the best solutions obtained during the execution of the algorithm in the given time is typically considered for applying a performance metric (Chicco & Mazza 2019; Liu et al. 2020).

When the constraints are imposed directly, it may happen that unfeasible solutions are generated during the calculations. These solutions have to be skipped or eliminated from the search. In this case, less useful solutions will be found by running the solver under analysis for a given number of times, worsening the performance indicator. Similar considerations apply when the solutions to be compared are subject to further conditions, for example, to satisfy the *N*-1 security conditions, as requested in Wang et al. (2008) for a transmission expansion planning problem.

Indications on the comparison strategies are provided in Liu et al. (2020), basically identifying pairwise comparison between algorithms (with the variants one-plays-all, generally used to check a new algorithm, and all-play-all or "round-robin"), and multi-algorithm comparison, both used in many contexts. For multi-algorithm comparison, statistical aggregations such as the cumulative distribution function are often used.

The comparison methods can be partitioned into static (with evaluation of the best solution, mean, standard deviation or other statistic outcomes), dynamic ranking (which considers the succession of the best values or static rankings during the time), and the cumulative distribution functions (considered at different times during the solution process). The latter type of comparison has become more and more interesting, also representing the confidence intervals (Liu et al. 2017; Doerr et al. 2018).

Liu et al. (2020) defined the problem of finding the best algorithm as a voting system, in which the algorithms are the candidates, the problems are the votes, and an algorithm performs better than the others if it exhibits better performance on more problems. However, they found the existence of the so-called "cycle ranking" or Condorcet paradox, namely, it may happen that different algorithms are winners for different problems, and it is not possible to conclude which algorithm is better overall. In practice, taking three algorithms A, B and C, it may happen that for different problems A is better than B, B is better than C, and C is better than A. The same concept is shown in Chicco & Mazza (2019) by indicating that the relation between the solvers is non-transitive, namely, if algorithm A is better than algorithm B for some problems, and algorithm B is better than algorithm C, this does not imply that algorithm A is better than algorithm C. Another paradox shown in Liu et al. (2020) is the so-called "survival of the fittest". In this case, the winner can be different by using different comparison strategies. The probability of occurrence of the two paradoxes is calculated based on the NFL assumption.

*4.3. Which superiority?*

Superiority is the term widely used to indicate that a given algorithm performs better than others. However, the way to assess superiority is often stated in a trivial and misleading way. In particular, the use of simple performance indicators such as the best solution, the average value of the solutions, or the standard deviation of the solutions, makes it possible to exacerbate the paradoxes indicated in the above section. The main reason is the lack of robustness of these indicators, especially the ones based on a single occurrence (such as the best value) that could be found occasionally during the execution of the algorithm (or even with a "lucky" choice of the initial population). The continuous production of articles claiming that the algorithm used is superior with respect to a selected set of other algorithms is mostly due to the use of these simple performance indicators. A synthesis of the mechanism that leads to this continuous production of articles has been provided in Chicco & Mazza (2019), by introducing a *perpetual motion* conceptual scheme, from which it is clear that it is not possible to find a formal and rigorous way to stop the production of articles.

The only way to reduce the number of articles with questionable superiority is to introduce more robust statistics-based indicators for comparing the algorithms with each other. A number of non-parametric statistical tests are summarized in Derrac et al. (2011). Another example is the *OPISD* indicator provided in Chicco & Mazza (2019), by considering the first-order stochastic dominance concepts (Hadar & Russell 1971) with the approach indicated in Chicco & Mazza (2017). Starting from the CDFs of the solutions obtained from a set of algorithms run on the same problem (a qualitative example with 3 algorithms is shown in Figure 4), the *OPISD* indicator is formulated by considering a reference CDF together with the CDFs obtained from the given algorithms, calculating for each algorithm the area $A$ between the corresponding CDF and the reference CDF (Figure 5). Then, the indicator is defined as $OPISD = (1+A)^{-1}$. In this way, the algorithm with the smallest area is the one that exhibits better performance. From Figure 5, the algorithm 2 is the one that exhibits the best performance.

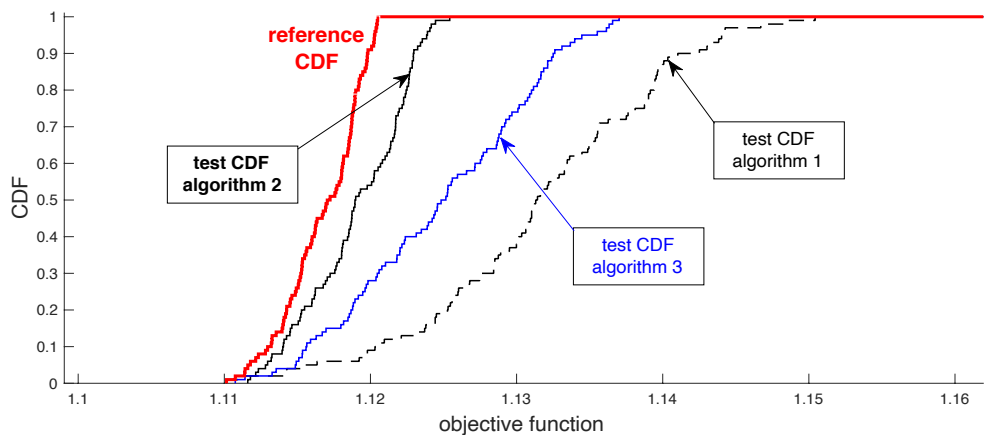

Figure 4. Determination of the reference CDF for the calculation of the *OPISD* indicator without knowing the global optimum.

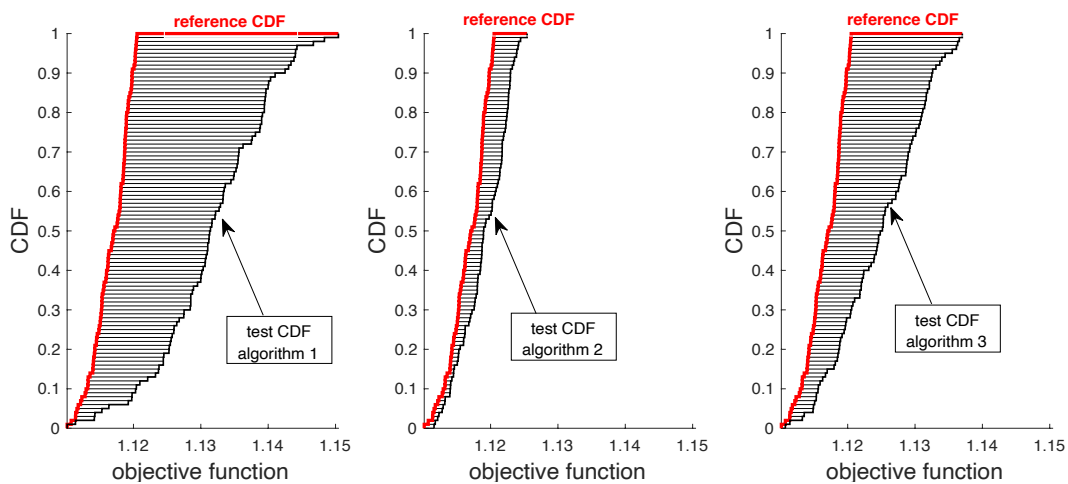

Figure 5. Determination of the areas for the calculation of the *OPISD* indicator.

The reference CDF is constructed in different ways, depending on whether the global optimum is known or not. If the global optimum is known, the reference CDF is equal to zero for values lower than the global optimum, and then it jumps to unity at the global optimum. This enables absolute comparisons among the algorithms, even though the global optimum can be known only in a few cases of relatively small systems, and "good" algorithms should always reach the global optimum for these small systems. If the global optimum is not known, the reference CDF is determined by starting from a given number of best solutions obtained from any of the algorithms used for comparison. In this case, only a relative comparison on the set of algorithms under analysis is possible, as the reference CDF changes each time.

## 5. Hybridization of the metaheuristics

The various metaheuristics have advantages and disadvantages, usually analyzed in terms of exploration and exploitation characteristics (Yang et al. 2014), and of contributions to improve the local search. To enhance the performance of the algorithms, one of the ways has been the formulation of hybrid optimization methods. The main types of hybridizations can be summarized as follows:
a) combinations of different heuristics; and,
b) combinations of metaheuristics with exact methods.

Successful strategies have been found from the combined use of a heuristic that carries out an extensive search in the solution space, together with a method suitable for local search. A practical example is the Evolutionary Particle Swarm Optimization (EPSO), in which an evolutionary model is used together with a particle movement operator to formulate a *self-adaptive* algorithm (Miranda & Fonseca, 2002). Another useful tool is the Lévy flights (Gutowski 2001), used to mitigate the issue of early convergence of metaheuristics (Yang & Deb 2009; Zhang et al. 2020) and get a better balance between exploration and exploitation. A further example of hybridization is the Differential Evolutionary Particle Swarm Optimization Algorithm (Garcia-Guarin et al. 2019) – the winner of the smart grid competition at the IEEE Congress on Evolutionary Computation/The Genetic and Evolutionary Computation Conference in 2019.

Depending on the problem under analysis, a useful practice can be the combination of a metaheuristic aimed at providing a contribution to the global search, and of an exact method of proven effectiveness to perform a local search. In many other cases, hybridizations have no special meaning and could be only aimed at producing further articles that contribute to the 'rush to heuristics'.

## 6. Multi-objective formulations

Multi-objective or many-objective optimization (see Section I) consider multiple objectives, and the solutions obtained for the individual objectives are relevant when *conflicting* objectives appear. Optimization with conflicting objectives does not search only the optimal values of the individual objectives, but identifies the compromise solutions as feasible alternatives for decision making.

Figure 6 shows the concept of dominated solution for a case with two objective functions $f_1$ and $f_2$ to be minimized. More generally, Figure 7 reports some qualitative examples of locations of the dominated solutions when the objective functions have to be maximized or minimized. Moreover, the dominated solutions can be assigned different *levels of dominance*, for assisting their ranking when they are used within solution algorithms. Figure 8 shows an example with four levels of dominance (where the first level is the best-known Pareto front, some points of which could be located onto the *true* Pareto front). Fuzzy-based dominance degrees have also been defined in Benedict & Vasudevan (2005).

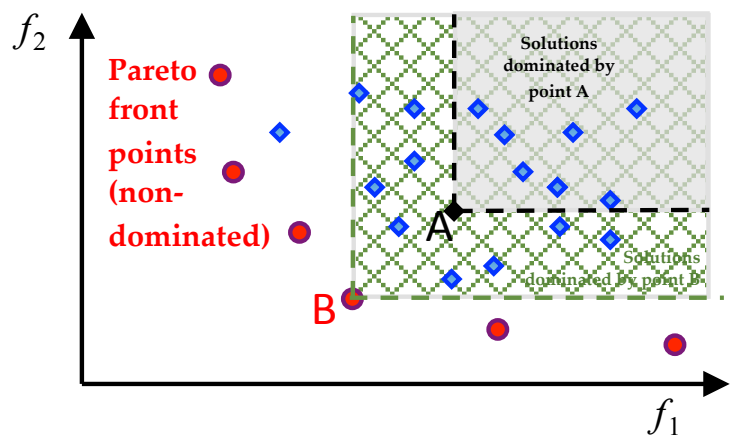

Figure 6. Concept of Pareto dominance. The functions $f_1$ and $f_2$ are minimized.

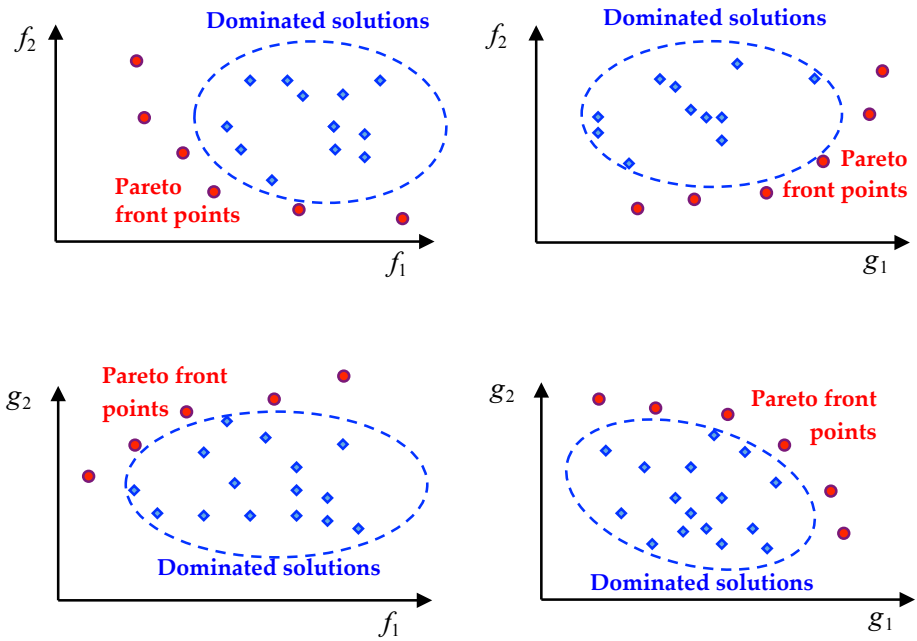

Figure 7. Pareto front points and dominated solutions for objective functions minimized ($f_1$ and $f_2$), or maximized ($g_1$ and $g_2$).

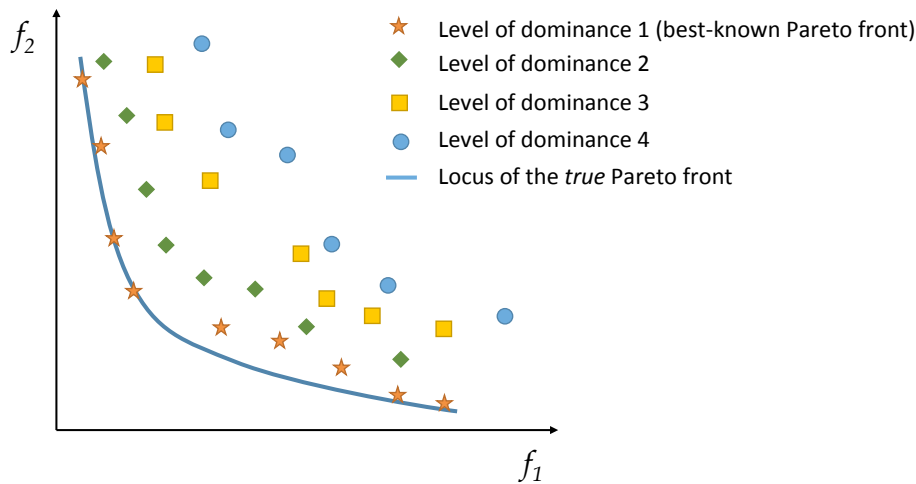

Figure 8. Levels of dominance. The functions $f_1$ and $f_2$ are minimized.

*6.1. Techniques for Pareto front calculation*

Some deterministic approaches are available. If the Pareto front is convex, the *weighted sum* of the individual objectives can be used to track the points of the Pareto front. Otherwise, other methods have to be exploited. The first type of possibility is to use the *ε-constrained method* (Haimes et al. 1971), which considers an individual objective as the target to be optimized and sets for all the other objectives a limit expressed by a threshold $\varepsilon$, then progressively reduces the threshold and upgrades the set of non-dominated solutions. Reduction to a single objective function is also carried out by using the goal programming approach (Contini 1968). *Fuzzy logic-based* approaches are also available (Deb 2001).

However, multi-objective optimization with Pareto front construction and assessment is a successful field of application of metaheuristics. In particular, the direct construction through metaheuristic approaches is an iterative process, in which at each iteration multiple solutions are generated and the solution set is then reduced to maintain only the non-dominated solutions. The dominated solutions are arranged into levels of dominance to enable wider comparisons.

Practically, for most of the many metaheuristic algorithms formulated for solving single-objective optimization problems, there is also the corresponding multi-objective optimization algorithm. As such, a list of metaheuristics for multi-objective optimization is not provided here. Only a few algorithms are mentioned because of their key historical and practical relevance: the Strength Pareto Evolutionary Approach (SPEA2, Zitzler et al. 2001), the Pareto Archived Evolution Strategy (PAES, Knowles & Corne 2000), and two versions of Non-dominated Sorting Genetic Algorithm, namely, NSGA-II (Deb et al. 2002) and NSGA-III (Deb & Jain 2014).

*6.2. No free lunches and comparisons among algorithms*

The discussion about the NFL theorem(s) is valid also for multi-objective optimization. Considering all the problems under analysis, according to the NFL theorem(s) all the algorithms outperform the other algorithms for some problems. However, for multi-objective optimization Corne & Knowles (2003) showed that the NFL does not generally apply when absolute performance metrics are used. This means that some multi-objective approaches can be better than others. On these bases, developing comparison metrics or quality indicators for multi-objective optimization algorithms is a challenging but worthwhile task. Some principles indicated in Zitzler et al. (2000) for the construction of effective multi-objective comparison metrics include:
a) The minimization of the distance between the best-known Pareto front and the true optimal Pareto front.
b) The presence of a distribution of the solutions as uniform as possible.
c) For each objective, the presence of a wide range of values in the best-known Pareto front.

The quality of each Pareto front obtained by using a multi-objective optimization algorithm is represented by using a real number. This number can be the average distance between the points located onto the Pareto front under analysis and the closest points of the best-known Pareto front. A survey of the indicators proposed in the literature is provided in Zitzler et al. (2003) and Zitzler et al. (2008). Other indicators are assessed with a chi-square-like deviation measure, to exploit the Pareto front diversity (Srinivas & Deb 1994; Zitzler et al. 2003). An Inverted Generational Distance indicator has been proposed in Bosman & Thierens (2003) to deal with the tradeoff between proximity and diversity preservation of the solutions in multi-objective optimization problems.

Furthermore, the hyper-volume indicator (Zitzler & Thiele 1999; Zitzler et al. 2003; While et al. 2006) has been used both for performance assessment and for guiding the search in various hyper-volume-based evolutionary optimizers (Augera et al. 2012). A weighted hyper-volume indicator has been introduced in (Zitzler et al. 2007), including weight functions to express user preferences. The selection of the weight functions and the transformation of the user preferences into weight functions has been addressed in (Brockhoff et al. 2013). While the idea of exploiting the hyper-volume calculation is interesting and based on geometric considerations, the determination of efficient algorithms to determine the hyper-volume when the number of dimensions increases is an open research field. For many objectives, a diversity metric has been proposed by Wang et al. (2017) by summing up the dissimilarity of the solutions to the rest of the population.

For power and energy systems, many multi-objective problems are defined with two or three objectives. In these cases, the hyper-volume can be calculated from available methods (Beume et al. 2009; Guerreiro & Fonseca 2018). In this case, it is also possible to extend previous results concerning the comparison among metaheuristics. Let us consider a number of objectives to be minimized. Following the concepts introduced in Chicco & Mazza (2019), when multiple metaheuristic algorithms have to be compared on a given multi-objective optimization problem, it is possible to determine the best-known Pareto front resulting from all the executions of the algorithm for a given time. Then, the comparison of all the Pareto fronts, resulting from the various methods, provide a quality indicator given by the hyper-volume included between the Pareto front under analysis and the best-known Pareto front. In this case, each solution is represented by using a scalar value. Lower values of this scalar value mean better quality of the result. The CDF of these scalar values can be constructed, and is considered as the reference CDF for *OPISD* calculation. The comparison between the CDFs of the individual metaheuristic algorithms and the reference CDF provides the area *A* to be used for *OPISD* calculation.

For comparing multi-objective metaheuristic algorithms, the definition of a suitable set of *test functions* is needed as a benchmark. For this purpose, classical test functions have been introduced in Zitzler et al. (2000), with the ZDT functions that contain two objectives, chosen to represent different cases with specific features. Further test functions have been introduced with nine DTLZ functions in Deb et al. (2001), as well as in Huband et al. (2006). However, for power and energy systems these benchmarks do not take into account the typical constraints that appear in specific problems, and as such an algorithm that shows good performance on these mathematical benchmarks could behave with poor performance on these problems. Lack of dedicated benchmarking for a wide set of power and energy problems does not enable the scientists in the power and energy domain presenting sufficiently broad results on the metaheuristic algorithm performance.

*6.3. Multi-objective solution ranking*

The last but not less important aspect concerning the multi-objective optimization outcomes is the possible *ranking* of the solutions determined by numerical calculations, to assist the decision-maker in the task of identifying the preferable solution. The methods available for this task require in some way to get the opinion of the expert to express preferences about the objectives considered. These methods belong to multi-criteria decision-making, where the criteria coincide with the objectives under consideration here. Some tools widely adopted are the Analytic Hierarchy Process (Saaty 1990), in which a 9-point scale quantifies the relative preferences between pairs of objectives, and the overall feasibility of the process is confirmed if an appropriately defined consistency criterion is satisfied. Furthermore, in the Ordered Weighted Averaging approach (Malczewski et al. 2003) the weights are ordered according to their relative importance, and a procedure driven by a single parameter is set up by using a transformation function that modifies the weighted values of the objectives. The Technique of Order Preference by Similarity to Ideal Solution (TOPSIS) method is based on the evaluation of the objectives depending on their distance to reference (ideal) points (Hwang & Yoon 1981; Mazza et al. 2012). Further methods such as ELECTRE (Roy 1968) and PROMETHEE (Brans & Mareschal 2005) are based on comparing pairs of weights. Other methods have been formulated by using fuzzy logic-based tools (Deb 2001; Benedict & Vasudevan 2005). For example, for a transmission expansion planning problem, the fuzzy logic-based tools are used in Moeini-Aghtaie et al. (2012), and in Jadidoleslam et al. (2017), where the rank of each solution is directly established in each Pareto front.

The reduction of personal judgment from the decision-maker is sometimes desired, especially when the problem is highly technical and the decision-maker has different qualifications. Moreover, in some cases, the introduction of automatic procedures to determine the relative importance of the objectives is needed, in particular when the judgment is included in an iterative process and the relative importance has to be established many times considering the variation of the objective function values.

These cases may typically occur when dealing with technical aspects in the power and energy systems domain. For example, a criterion to compare non-dominated solutions based on power systems concepts has been introduced in Maghouli et al. (2009), in which the Incremental Cost Benefit ratio has been defined by calculating the ratio between the congestion cost reduction with respect to the base case and the investment referring to the solution considered. The automatic creation of the entries of the pair comparison matrices for the AHP approach has been introduced in Mazza et al. (2014) for a distribution system reconfiguration problem, by using an affine function that maps the objective function values onto the Saaty interval from 1 to 9.

**7. Effectiveness of metaheuristic-based optimization: pitfalls and inappropriate statements**

In their articles aimed at applying metaheuristic optimization methods, various authors include inappropriate statements on the effectiveness of the methods used. These statements are also one of the main reason of the rejection of many papers sent to scientific journals or conferences. The most significant (and sometimes common) situations are recalled in this section, with the corresponding discussion on whether more appropriate solutions could be adopted.

*7.1. On reaching the global optimum*

Some articles report that the scope of the analysis is "to reach the global optimum". This statement is *never correct* for a heuristic run on a large-scale problem. In the previous sections, it has been clarified that no metaheuristic can guarantee to find the global optimum for any finite time or number of iterations. At most, asymptotic convergence to the global optimum has been proven for some metaheuristics, namely, by executing an infinite number of iterations. In practice, *no heuristic* is able to guarantee that the global optimum can be reached in a finite time, as it would be needed for engineering problems.

*7.2. Adaptive stop criterion*

As a consequence of the previous point, how to decide when to stop the execution of a metaheuristic algorithm becomes a crucial issue. Setting up a sound stop criterion (or termination criterion) is needed. Quite surprisingly, many algorithms used in available publications consider the maximum number of iterations $N_{max}$ as the sole stop criterion. However, this choice is generally inappropriate. In fact, two different issues could occur (Chicco & Mazza 2013):

1) *early stopping*, in which the execution could be stopped when the evolution of the objective *function is s*till providing *significant* improvements (Figure 9a); or

2) *late stopping*, in which the execution could be stopped when the solution had no variations (or no significant changes, in a milder version) for many of the last iterations (Figure 9b); in this case, the last part of the execution, with many constant values, is unnecessary and could have been avoided.

Improvements in the objective function(s) could appear at any time. However, the identification of a sound stop criterion is important, to use the computation time in the best way.

The definition of an *adaptive stop criterion* is the most appropriate solution to the above-indicated issues. In the adaptive stop criterion (sometimes indicated as *stagnation* criterion), the algorithm terminates when no change occurs in the best objective function found so far, after a given number $N_s$ of successive iterations of the algorithm (Figure 9c). In this case, both early stopping and late stopping are avoided. The number of successive iterations is a user-defined parameter that can also be chosen based on the experience on the variability of the objective function for specific problems. The maximum number of iterations, set to a very high value, could remain as a *last-resource* stop criterion to trap possible unlimited executions.

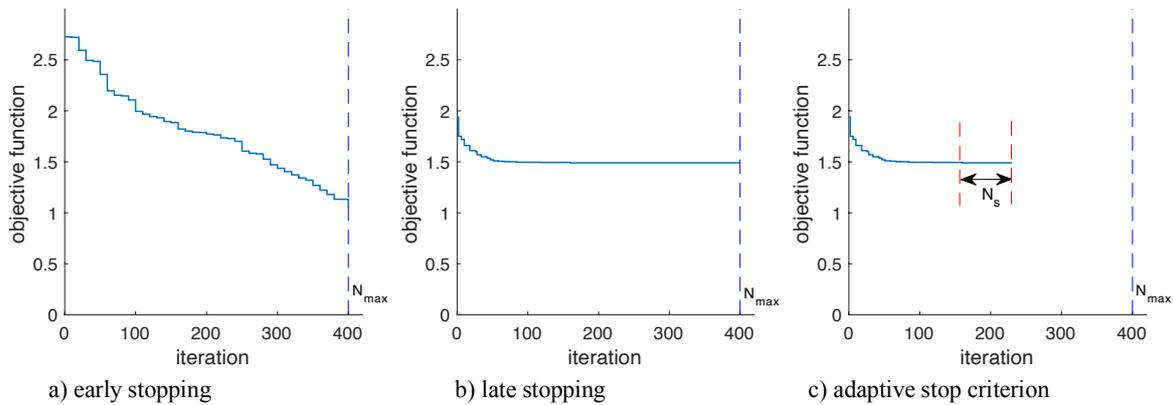

a) early stopping  b) late stopping  c) adaptive stop criterion

Figure 9. Effectiveness of the adaptive stop criterion for objective functions to be minimized. $N_{max}$: fixed maximum number of iterations. $N_s$: number of successive iterations without objective function improvements.

If the metaheuristic algorithm is run for performance comparisons (see Section 4.2), the stop criterion (equal for all algorithms) could become the computation time. Also in this case, the use of the maximum number of iterations is not needed. Thereby, the maximum number of iterations should not be used as the primary criterion to terminate the execution of a metaheuristic algorithm.

*7.3. Not only metaheuristics*

A common but inappropriate trend found in many papers is to use a metaheuristic method or variant and to compare it with other metaheuristics only. While this trend is highly questionable because of the 'rush to heuristics' issue (see Section 2), the availability of many alternative methods beyond metaheuristics has to be considered. Indeed, for specific problems, there can be many algorithms of different types, and the rationale for using a "new" metaheuristic has to be explicitly stated. It is not sufficient to compare a few algorithms chosen at random. In general, metaheuristic algorithms are relatively simple to be implemented (at least in their standard versions and for problems that do not require setting up equality or inequality constraints referring to complex structures). For this reason, it could be easier to perform comparisons among metaheuristics taken from existing libraries or implemented by the same authors. However, a fair comparison requires choosing a set of metaheuristics that have exhibited good performances for a number of related problems. The absence of well-established benchmarks leaves the situation somehow confused.

*7.4. The importance of fast convergence*

Some contributions give key importance to achieving *fast convergence* (in terms of *number of iterations* employed to provide the algorithm's result) and show the case with the fastest convergence as an example of the "superiority" of the proposed algorithm. An example of statement for this case is "The convergence of the proposed metaheuristic is better than for other methods". What in reality happens is that, during the iterative process, the objective function of the proposed metaheuristic improves in *fewer iterations* than for other methods tested.

First of all, the computation time for solving one iteration for a given algorithm is generally different with respect to other methods. As such, just considering the number of iterations is rather meaningless. Moreover, the improvements occurring in fewer iterations *could not mean anything*. In general, fast convergence can be achieved from good (i.e., lucky) choices of the initial case for single-update methods, or of one or more individuals of the initial population in population-based methods. In the extreme case, if the initial choice happens to contain the global optimum (without knowing it neither in advance, nor having any way to be sure of that), the convergence to the best solution is immediate. However, by no way, this could mean that the algorithm used is better than another one.

*7.5. How many executions?*

A typical drawback of many articles is the limited number of executions performed with the metaheuristic method. In order to reach statistical significance for the approach used, the number of executions has to be sufficiently high, indicatively not less than 100. However, if the problem can be solved thousands of times within a reasonable computation time, it would be even better. As an obvious corollary, if the iterative process has been run only *once* for each method, a comparison among *one-shot* cases is not significant to reach any conclusion.

When a comparison based on a given computation time limit is carried out (e.g., within a competition), the number of executions will be driven by the computation time limit. In that case, writing an efficient and fast programming code would be clearly relevant.

*7.6. Avoiding fallacies*

During the testing of a metaheuristic algorithm on a well-known problem, a possible issue is that "the results obtained are impressively better than those appearing in the literature". What could happen in this case is that the proposed heuristic has provided largely better results with respect to the same or similar solutions found in other literature references from various other methods run on the same problem. The warning for such a case is that this kind of result may be *suspect*. The possible causes of this situation have to be searched on possible issues in the modeling used. For example, the system considered could be not exactly the same as in the other references. Another possibility is that the constraints may have some differences among the methods compared (in terms of number, formulation, or threshold values applied). A possible hint, in this case, is to solve the optimization on a small system for which the global optimum is known: if the results differ, there could be something wrong in the data used or in the implementation of the algorithm.

# 8. Conclusions

This paper has presented a contribution to the discussion on the use of metaheuristic methods in the solution of global optimization problems. Starting from the evidence that the number of published articles on metaheuristic applications in the power and energy systems domain is increasing with an impressively high trend, some questions on the reasons of this 'rush to heuristics' have been addressed. It has emerged that there is a lack of dedicated benchmark for global optimization in the power and energy systems domain, as well as a lack of statistically significant and robust comparisons among the outcomes of the metaheuristic approach. The existing benchmarks used in the evolutionary computation domain are not always sufficient to represent the specific aspects encountered in power and energy systems problems. Therefore, dedicated customizations may be needed to execute some metaheuristic algorithms on these problems. The construction of specific benchmarks for given problems in the power and energy systems area is a challenging topic for future research. The basic literature concerning comparison metrics for single-objective and multi-objective problems solved with a metaheuristic approach has been reviewed. In many articles, the metrics used for comparisons are too weak to reach sound conclusions on the effectiveness of the proposed algorithm. Sometimes, the superior performance of the method used over a selected set of other methods is incorrectly claimed on the basis of a few results. A set of underlying principles has been identified, to explain the characteristics of the metaheuristic algorithms in a systematic way and discover possible similarities among the many algorithms proposed.

The discussion on the most effective use of metaheuristic algorithms is a challenging subject. Some pitfalls and inappropriate statements, sometimes found in literature contributions, have been highlighted. While there is no formal way to stop the proliferation of articles that propose new applications, variants or hybridizations of metaheuristics, it is the authors' idea that systematic indications on how to avoid these pitfalls may be useful for the researchers. For this purpose, some *guidelines* for preparing sound contributions on the application of metaheuristic algorithms to power and energy system problems (but also useful for other application fields) are summarized in the following points:

1) Consider the size and type of the *optimization* problem. If the problem can be solved with exhaustive search or by using exact tools in reasonable computation time, then applying a metaheuristic algorithm is useless.
2) For *proposing* a new metaheuristic method or variant referring to the mechanism of the method (including hybridizations) and not to direct customization to specific needs of problems in the power and energy area, send the contribution to journals referring to soft computing and evolutionary computation domains, where a conceptual and practical validation can be carried out by specialists.
3) Specify the *information coding* in details.
4) Clarify the *relations* between the heuristic operators and the variables of the specific problem (not only describing general-purpose tools).
5) Illustrate the treatment of the *constraints* explicitly. Customization of the classical version of a metaheuristic algorithm could be needed, and the rationale and effectiveness of the customization have to be specifically addressed. Discuss how to keep the *constraints* enforced during the evolution of the computational process.
6) Explain *parameter settings* and values, possibly carrying out *sensitivity* analyses.
7) Choose the algorithms to compare for obtaining a reasonably strong *benchmark* (not only selecting a few other metaheuristics for which relevant results have not been clearly stated – with an accurate look at the state of the art). Avoid the mere 'rush to heuristics'! General benchmarks defined with mathematical test functions could not be detailed enough for representing the specific issues that appear in power and energy systems domain.
8) Implement the *adaptive stop criterion* (using the maximum number of iterations only as a secondary criterion to terminate the execution).
9) Implement the algorithms to be compared and execute them with the *same data and problem formulation*, to avoid possible variations with respect to the data and problem definition used in other articles.
10) Show the *statistics* of the results obtained on test systems and/or real networks, based on a significant number of executions (not only one execution) and on the use of appropriately robust statistical indicators.
11) Use the correct *terminology*, avoiding to declare the superiority of an algorithm on the basis of the results obtained on a specific problem only and with limited testing.

The previous indications refer to the application side and are directed to the scientific communities that adopt metaheuristics for solving problems that are not solvable with exact methods, or that can be solved efficiently with metaheuristic algorithms. However, the concept of "efficient" solution has not been clearly explained yet. Substantial work is needed and is in progress in the evolutionary computation community, in the direction of improving the design of metaheuristics and developing modeling languages and efficient general-purpose solvers (Sörensen et al. 2018). This direction does not include the 'rush to heuristics', which is just wasting a lot of energies in a useless and non-concluding 'perpetual motion' of production of contributions with improper developments and incorrect attempts to declare an inexistent 'superiority'.

# References

Andrei, H.; Chicco, G. Identification of the radial configurations extracted from the weakly meshed structures of electrical distribution systems. IEEE Trans. on Circuits and Systems I, Reg. Papers, 2008;55(4):1149–1158.

Augera, A.; Bader, J.; Brockhoff, D.; Zitzler, E. Hypervolume-based multiobjective optimization: Theoretical foundations and practical implications. Theoretical Computer Science, 2012;425:75–103.

Bäck, T. Selective pressure in evolutionary algorithms: A characterization of selection mechanisms. Proceedings of the 1st Conference on Evolutionary Computing, 1994: 57–62.

Bain, S.; Thornton, J.: Sattar, A. Methods of automatic algorithm generation. PRICAI 2004: Trends in Artificial Intelligence, Springer, 2004: 144–153.

Bartz-Beielstein, T.; Chiarandini, M.; Paquete, L.; Preuss, M. Experimental Methods for the Analysis of Optimization Algorithms. Springer, Berlin Heidelberg, 2010.

Batrinu, F.; Carpaneto, E.; Chicco, G. A unified scheme for testing alternative techniques for distribution system minimum loss reconfiguration. Proc. International Conference on Future Power Systems - FPS 2005, Amsterdam, The Netherlands, November 16-18, 2005, paper O11-09.

Bélisle, C.J.P. Convergence theorems for a class of simulated annealing algorithms on $R_d$. J. Appl. Probab., 1992;29(4):885–895.

Benedict, S.; Vasudevan, V. Fuzzy-Pareto-dominance and its application in evolutionary multi-objective optimization. Proc. 3rd Int. Conf. Evol. Multi-Criterion Optim (EMO), 2005, Berlin, Germany: 399–412.

Beume, N.; Fonseca, C.M.; Lopez-Ibanez, M.; Paquete, L.; Vahrenhold, J. On the Complexity of Computing the Hypervolume Indicator. IEEE Trans. on Evolutionary Computation, 2009;13(5):1075–1082.

Blum, C.; Dorigo, M.; The hyper-cube framework for ant colony optimization. IEEE Transactions on Systems, Man and Cybernetics: Part B, 2004;34(2):1161–1172.

Bosman, P.; Thierens D. The balance between proximity and diversity in multiobjective evolutionary algorithms. IEEE Trans. Evol. Comput., 2003;7(2):174–188.

Boussaïd, I.; Lepagnot, J.; Siarry, P. A survey on optimization metaheuristics. Information Sciences, 2013;237:82–117.

Brans, J.P.; Mareschal, B. Promethee Methods. In Multiple Criteria Decision Analysis: State of the Art Surveys, International Series in Operations Research & Management Science, Springer, New York, NY, 2005;78:163–195.

Brockhoff, D.; Bader, J.; Thiele, L.; Zitzler, E. Directed Multiobjective Optimization Based on the Weighted Hypervolume Indicator. J. Multi-Crit. Decis. Anal., 2013;20:291–317.

Burke, E.K.; Hyde, M.R. ; Kendall, G.; Ochoa, G.; Özcan, E. A classification of hyper-heuristic approaches. In Handbook of Metaheuristics (International Series in Operations Research & Management Science), vol. 146, 2nd ed., Gendreau, M. & Potvin, J.-Y. (Eds). Springer, New York, NY, USA, 2010:449–468.

Campos, V.E.M.; Pereira, A.G.C.; Rojas Cruz, J.A. Modeling the genetic algorithm by a non-homogeneous Markov chain: weak and strong ergodicity. Theory of Probability and its Applications, 2013;57(1):144–151.

Carpaneto, E.; Chicco, G. Distribution system minimum loss reconfiguration in the Hyper-Cube Ant Colony Optimization framework. Electric Power Systems Research, 2008;78(12):2037–2045.

Carreno, E.M.; Romero, R.; Padilha-Feltrin, A. An Efficient Codification to Solve Distribution Network Reconfiguration for Loss Reduction Problem. IEEE Transactions on Power Systems, 2008;23:1542–1551.

Cerf, R.. Asymptotic convergence of genetic algorithms. Advances in Applied Probability, 1998;30:521–550.

Chen, G.; Low, C.P.; Yang, Z. Preserving and Exploiting Genetic Diversity in Evolutionary Programming Algorithms. IEEE Transactions on Evolutionary Computation, 2009;13:661–673.

Chicco, G.; Mazza, A. An Overview of the Probability-based Methods for Optimal Electrical Distribution System Reconfiguration. Proc. 4th International Symposium on Electrical and Electronics Engineering (ISEEE), Galati, Romania, 10-12 October 2013.

Chicco, G.; Mazza, A. Assessment of Optimal Distribution Network Reconfiguration Results using Stochastic Dominance Concepts. Sustainable Energy, Grids and Networks, 2017;9:75–79.

Chicco, G.; Mazza, A. Heuristic Optimization of Electrical Energy Systems: Refined Metrics to Compare the Solutions. Sustainable Energy, Grids and Networks, 2019;17:100197.

Contini, B. A stochastic approach to goal programming. Operations Research, 1968;16(3):576–586.

Corne, D.W.; Knowles, J.D. No Free Lunch and Free Leftovers Theorems for Multiobjective Optimisation Problems. In: Fonseca C.M.; Fleming P.J.; Zitzler E.; Thiele L.; Deb K. (eds) Evolutionary Multi-Criterion Optimization, EMO 2003. Lecture Notes in Computer Science, Springer, Berlin, Heidelberg, 2003;2632.

Črepinšek, M.; Liu, S.H.; Mernik, M. Exploration and exploitation in evolutionary algorithms: A survey. ACM Computing Surveys (CSUR), 2013;45:1–33.

Deb, K. Multi-Objective Optimization Using Evolutionary Algorithms, Wiley, New York, 2001.

Deb, K.; Thiele, L.; Laumanns, M.; Zitzler, E. Scalable Test Problems for Evolutionary Multi-Objective Optimization. Kanpur, India: Kanpur Genetic Algorithms Lab. (KanGAL), Indian Inst. Technol., 2001. KanGAL Report 2 001 001.

Deb, K.; Pratap, A.; Agarwal, S.; Meyarivan, T. A fast and elitist multiobjective genetic algorithm: NSGA-II. IEEE Trans. Evol. Comput., 2002;6:182–197.

Deb, K.; Jain, H. An evolutionary many-objective optimization algorithm using reference-point-based nondominated sorting approach, part I: Solving problems with box constraints. IEEE Trans. Evol. Comput., 2014;18(4):577–601.

del Valle, Y.; Venayagamoorthy, G.K.; Mohagheghi, S.; Hernandez, J.-C.; Harley, R.G. Particle swarm optimization: Basic concepts, variants and applications in power systems. IEEE Transactions on Evolutionary Computation, 2008;12:171–195.

Derrac, J.; García, S.; Molina, D.; Herrera, F. A practical tutorial on the use of nonparametric statistical tests as a methodology for comparing evolutionary and swarm intelligence algorithms. Swarm Evolut. Comput., 2011;1(1):3–18.

Doerr, C.; Wang, H.; Ye, F.; van Rijn, S.; Bäck, T. IOHprofiler: A Benchmarking and Profiling Tool for Iterative Optimization Heuristics. arXiv e-prints:1810.05281, Oct. 2018. [Online]. Available: https://arxiv.org/abs/1810.05281.

Dolan, E.D.; Moré, J.J. Benchmarking optimization software with performance profiles. Math. Program., 2002;91(2):201–213.

Dokeroglu, T.; Sevinc, E.; Kucukyilmaz, T.; Cosar, A. A survey on new generation metaheuristic algorithms. Computers & Industrial Engineering, 2019;137:106040.

Dorigo, M.; Maniezzo, V.; Colorni, A. Positive Feedback as a Search Strategy. Politecnico di Milano: Dipartimento di Elettronica, 1991.

Drake, J.H.; Kheiri, A.; Özcan, E.; Burke, E.K. Recent advances in selection hyper-heuristics. European Journal of Operational Research, 2020;285(2):405-428.

Eiben, A.E.; Aarts, E.H.L.; Van Hee, K.M. Global convergence of genetic algorithms: a Markov chain analysis. In: Schewefel, H.P.; Männer, R. (Eds.), Parallel Problem Solving from Nature. Springer, Berlin, Heidelberg, 1991: 4–12.

Garcia-Guarin, J.; Rodriguez, D.; Alvarez, D.; Rivera, S.; Cortes, C.; Guzman, A.; Bretas, A.; Aguero, J.R.; Bretas, N. Smart Microgrids Operation Considering a Variable Neighborhood Search: The Differential Evolutionary Particle Swarm Optimization Algorithm. Energies 2019;12:3149.

Gaviano, M.; Kvasov, D.; Lera, D.; Sergeyev, Y.D. Algorithm 829: Software for generation of classes of test functions with known local and global minima for global optimization. ACM Transactions on Mathematical Software, 2003;9:469–480.

Giraud-Carrier, C.; Provost, F. Toward a justification of meta-learning: Is the no free lunch theorem a show-stopper. Proceedings of the ICML-2005 Workshop on Meta-learning, 2005:12–19.

Guerreiro, A.P.; Fonseca, C.M. Computing and Updating Hypervolume Contributions in Up to Four Dimensions. IEEE Transactions on Evolutionary Computation, 2018;22(3):449–463.

Gutowski, M. Levy flights as an underlying mechanism for global optimization algorithms. ArXiv Mathematical Physics e-Prints, June 2001.

Hadar, J.; Russell, W.R. Stochastic dominance diversification. J. Econom. Theory 1971;3:288–305.

Haimes, Y.; Lasdon, L.; Wismer, D. On a bicriterion formulation of the problems of integrated system identification and system optimization. IEEE Transactions on Systems, Man, and Cybernetics, 1971;1:296–297.

Hansen, N.; Auger, A.; Ros, R.; Finck, S.; Posik, P. Comparing results of 31 algorithms from the black-box optimization benchmarking bbob-2009. Proceedings of the 12th annual conference companion on Genetic and evolutionary computation, 2010:1689–1696.

Holland, J.H. Outline for a logical theory of adaptive systems. Journal of the Association for Computing Machinery, 1962;9(3):297–314.

Holland, J.H. Adaptation in Natural and Artificial Systems. The University of Michigan Press, Ann Arbor, MI, 1975.

Huband, S.; Hingston, P.; Barone, L.; While, L. A review of multiobjective test problems and a scalable test problem toolkit. IEEE Transactions on Evolutionary Computation, 2006;10(5):477–506.

Hwang, C.L.; Yoon, K. Multiple attribute decision making. Methods and applications: a state-of-the-art survey. Springer-Verlag, Berlin and New York, 1981.

Ishibuchi, H.; Tsukamoto, N.; Nojima, Y. Evolutionary many-objective optimization: A short review. Proc. 2008 IEEE Congr. Evol. Comput. (CEC), Hong Kong, 2008:2419–2426.

Jadidoleslam, M.; Ebrahimi, A.; Latify, M.A. Probabilistic transmission expansion planning to maximize the integration of wind power. Renewable Energy, 2017;114(B):866–878.

Kazarlis, S.A.; Bakirtzis, A.G.; Petridis, V. A genetic algorithm solution to the unit commitment problem. IEEE Transactions on Power Systems, 1996;11(1):83–92.

Knowles, J.D.; Corne, D.W. Approximating the nondominated front using the Pareto archived evolution strategy. Evol. Comput., 2000;8(2):149–172.

Koza, J.R. Genetic programming II: Automatic discovery of reusable subprograms. Cambridge, MA, USA, 1994.

Kirkpatrick, S.; Gelatt, C.D.; Vecchi, M.P. Optimization by simulated annealing. Science, 1983;220:671–680.

Lee, K.Y.; El-Sharkawi, M.A. (Eds.). Modern Heuristic Optimization Techniques, Wiley, Hoboken, NJ, 2008.

Li, K.; Deb, K.; Zhang, Q.; Kwong, S. An Evolutionary Many-Objective Optimization Algorithm Based on Dominance and Decomposition. IEEE Transactions on Evolutionary Computation, 2015;19(5);694–716.

Liang, J.J.; Qu, B.Y.; Suganthan, P.N. Problem definitions and evaluation criteria for the CEC 2013 special session and competition on real-parameter optimization, Computational Intelligence Laboratory, Zhengzhou University, Zhengzhou China and Nanyang Technological University, Singapore, Tech. Rep. 201212, January 2013.

Liu, Q.; Chen, W.N.; Deng, J.D.; Gu, T.; Zhang, H.; Yu, Z.; Zhang, J. Benchmarking stochastic algorithms for global optimization problems by visualizing confidence intervals. IEEE Transactions on Cybernetics, 2017;47:2924–2937.

Liu, Q.; Gehrlein, W.V.; Wang, L.; Yan, Y.; Cao, Y.; Chen, W.; Li, Y. Paradoxes in Numerical Comparison of Optimization Algorithms. IEEE Transactions on Evolutionary Computation, 2020, in press.

Locatelli, M.; Convergence properties of simulated annealing for continuous global optimization. J. Appl. Probab. 1996;33:1127–1140.

Maghouli, P.; Hosseini, S.H.; Buygi, M.O.; Shahidehpour, M. A Multi-Objective Framework for Transmission Expansion Planning in Deregulated Environments. IEEE Trans. Power Syst., 2009;24(2):1051–1061.

Malczewski, J.; Chapman, T.; Flegel, C.; Walters, D.; Shrubsole, D.; Healy, M.A. GIS–multicriteria evaluation with ordered weighted averaging (OWA): case study of developing watershed management strategies. Environment and Planning A, 2003;35:1769–1784.

Mazza, A.; Chicco, G. Application of TOPSIS in distribution systems multi-objective optimization. Proc. 9th World Energy System Conference, Suceava, Romania, 28-30 June 2012, pp. 625–633.

Mazza, A.; Chicco, G.; Russo, A. Optimal multi-objective distribution system reconfiguration with multi criteria decision making-based solution ranking and enhanced genetic operators. Electrical Power & Energy Systems, 2014;54;255–267.

Mersmann, O.; Preuss, M.; Trautmann, H.; Bischl, B.; Weihs, C. Analyzing the BBOB results by means of benchmarking concepts. Evolutionary Computation, 2015;23:161–185.

Miranda, V.; Fonseca, N. EPSO - best-of-two-worlds meta-heuristic applied to power system problems. Proceedings of the 2002 Congress on Evolutionary Computation (CEC'02); 2:1080–1085.

Mitsos, A.; Najman, J.; Kevrekidis, I.G. Optimal deterministic algorithm generation. J Glob Optim, 2018;71:891–913.

Moeini-Aghtaie, M.; Abbaspour, A.; Fotuhi-Firuzabad, M. Incorporating large-scale distant wind farms in probabilistic transmission expansion planning; Part I: theory and algorithm. IEEE Trans. Power Syst., 2012;27:1585–1593.

Moré, J.J.; Wild, S.M. Benchmarking derivative-free optimization algorithms. SIAM J. Optim., 2009;20(1);172–191.

Muralikrishnan, N.; Jebaraj, L.; Rajan, C.C.A. A Comprehensive Review on Evolutionary Optimization Techniques Applied for Unit Commitment Problem. IEEE Access, 2020;8: 132980–133014.

Pardalos, P.M; Edwin Romeijn, H.; Tuy, H. Recent developments and trends in global optimization, Journal of Computational and Applied Mathematics, 2000;124(1–2):209–228.

H.E. Romeijn, R.L. Smith, Simulated annealing for constrained global optimization, J. Global Optim., 1994;5:101–126.

Rojas Cruz, J.A.; Pereira, A.G.C. The elitist non-homogeneous geneticalgorithm: Almost sure convergence. Statistics and Probability Letters, 2013;83:2179–2185.

Roy, B. Classement et choix en présence de points de vue multiples, (in French). Revue française d'informatique et de recherche opérationnelle, 1968;2(V1):57–75.

Rudolph, G. Convergence analysis of canonical genetic algorithms. IEEE Transactions on Neural Networks, 1994;5:96–101.

Rudolph, G. On a multi-objective evolutionary algorithm and its convergence to the Pareto set. Proceedings of the 5th IEEE Conference on Evolutionary Computation, 1998. IEEE Press, Piscataway, pp. 511–516.

Rudolph, G.; Agapie, A. Convergence properties of some multi-objective evolutionary algorithms. Proceedings of the 2000 Conference on Evolutionary Computation. IEEE Press, Piscataway, 2000;2:1010–1016.

Saaty, T.L. How to make a decision: the analytic hierarchy process. European Journal of Operational Research, 1990;48:9–26.

Salcedo-Sanz, S. Modern meta-heuristics based on nonlinear physics processes: A review of models and design procedures. Physics Reports, 2016;655:1–70.

Sörensen, K. Metaheuristics—the metaphor exposed. International Transactions in Operational Research, 2015;22(1):3–18.

Sörensen K., Sevaux M., Glover F. A History of Metaheuristics. In Martí, R.; Pardalos, P.; Resende, M. (Eds), Handbook of Heuristics. Springer, Cham, Switzerland, 2018.

Srinivas, N., Deb, K. Multiobjective optimization using nondominated sorting in genetic algorithms, Evolutionary Computation, 1994;2(3):221–248.

Swarup, K.S, Yamashiro, S. Unit commitment solution methodology using genetic algorithm. IEEE Transactions on Power Systems, 2002;17(1):87–91.

Taillard, É.D.; Gambardella, L.M.; Gendreau, M.; Potvin, J.Y. Adaptive memory programming: a unified view of metaheuristics. European Journal of Operational Research, 2001;135:1–16.

Tejada-Arango, D.A.; Lumbreras, S.; Sánchez-Martín, P.; Ramos, A. Which Unit-Commitment Formulation is Best? A Comparison Framework. IEEE Transactions on Power Systems, 2020;35(4):2926–2936.

Tomoiaga, B.; Chindris, M.; Sumper, A.; Sudria-Andreu, A.; Villafafila-Robles, R. Pareto optimal reconfiguration of power distribution systems using a genetic algorithm based on NSGA-II. Energies, 2013;6:1439–1455.

Villalobos-Arias, M., Coello, C.A.C., Hernández-Lerma, O. Asymptotic Convergence of Some Metaheuristics Used for Multiobjective Optimization. In Wright A.H., Vose M.D., De Jong K.A., Schmitt L.M. (eds) Foundations of Genetic Algorithms, FOGA 2005. Lecture Notes in Computer Science, vol. 3469, 2005. Springer, Berlin, Heidelberg.

Villalobos-Arias, M.; Coello Coello, C.A.; Hernández-Lerma, O. Asymptotic convergence of a simulated annealing algorithm for multiobjective optimization problems. Math. Meth. Oper. Res., 2006;64:353–362.

Wang, Y.; Cheng, H.; Wang. C.; Hu, Z.; Yao, L.; Ma, Z.; Zhu, Z. Pareto Optimality-based Multi-objective Transmission Planning Considering Transmission Congestion. Electric Power System Research, 2008:78(9):1619–1626.

Wang, H.; Jin, Y.; Yao, X. Diversity assessment in many-objective optimization. IEEE Trans. Cybern., 2017;47(6):1510–1522.

While, R.L.; Hingston, P.; Barone, L.; Huband, S. A faster algorithm for calculating hypervolume. IEEE Trans. Evol. Comput., 2006;10(1):29–38.

Wolpert, D.H., Macready, W.G. No Free Lunch Theorems for Optimization, IEEE Transactions on Evolutionary Computation, 1997;1:67–82.

Wolpert, D.H.; Macready, W.G. Coevolutionary free lunches. IEEE Transactions on Evolutionary Computation, 2005;9(6):721–735.

Yang, X.S.; Deb, S. Cuckoo Search via Lévy flights. 2009 World Congress on Nature & Biologically Inspired Computing (NaBIC); 2009:210–214.

Yang, X.S.; Deb, S.; Fong, S. Metaheuristic Algorithms: Optimal Balance of Intensification and Diversification. Applied Mathematics & Information Sciences Journal, 2014:8(3):977–983.

Zedadra, O.; Guerrieri, A.; Jouandeau, N.; Spezzano, G.; Seridi, H.; Fortino, G. Swarm intelligence-based algorithms within IoT-based systems: A review. Journal of Parallel and Distributed Computing, 2018;122:173–187.

Zhang, X.; Xu, Y.; Yu, C.; Heidari, A.A.; Li, S.; Chen, H.; Li, C. Gaussian mutational chaotic fruit fly-built optimization and feature selection. Expert Systems with Applications, 2020;141:112976.

Zheng, Q.P.; Wang, J.; Liu, A.L. Stochastic Optimization for Unit Commitment—A Review. IEEE Transactions on Power Systems, 2015;30(4):1913–1924.

Zitzler, E.; Thiele, L. Multiobjective evolutionary algorithms: A comparative case study and the strength Pareto approach. IEEE Trans. Evol. Comput., 1999;3(4):257–271.

Zitzler, E.; Deb, K.; Thiele, L. Comparison of Multiobjective Evolutionary Algorithms: Empirical Results. Evolutionary Computation, 2000;8(2):173–195.

Zitzler, E.; Laumanns, M.; Thiele, L. SPEA2: Improving the Strength Pareto Evolutionary Algorithm. TIK-report, 103, 2001.

Zitzler, E.; Thiele, L.; Laumanns, M.; Fonseca, C.M.; Grunert da Fonseca, V. Performance assessment of multiobjective optimizers: an analysis and review. IEEE Trans. on Evolutionary Computation, 2003;7(2):117–132.

Zitzler, E.; Brockhoff, D.; Thiele, L. The hypervolume indicator revisited: on the design of Pareto-compliant indicators via weighted integration. Proc. of Conference on Evolutionary Multi-Criterion Optimization (EMO 2007), 2007:862–876.

Zitzler, E.; Knowles, J.; Thiele, L. Quality assessment of Pareto set approximations. In Branke, J.; Deb, K.; Miettinen, K.; Slowinski, R. (Eds) Multiobjective Optimization: Interactive and Evolutionary Approaches, Springer, Berlin Heidelberg, 2008.